\documentclass{article}
\usepackage{hyphenat}
\usepackage{microtype}
\usepackage{graphicx}
\graphicspath{{figures/}{figures/qualitative/}{sec/}} 
\usepackage{subcaption}
\usepackage{booktabs} 
\usepackage{multirow}
\usepackage{longtable}
\usepackage{float}
\usepackage{caption}

\usepackage{hyperref}
\usepackage[ruled,vlined]{algorithm2e}

\usepackage{wrapfig}\usepackage[table]{xcolor} 
\definecolor{bestcolor}{RGB}{219, 208, 237}
\definecolor{secondcolor}{RGB}{241, 237, 248}
\definecolor{line-blue}{RGB}{243, 248, 252}
\definecolor{taskSpatial}{RGB}{255, 204, 170}   
\definecolor{taskTransform}{RGB}{210, 190, 230} 
\definecolor{taskLogic}{RGB}{160, 200, 255}     
\definecolor{taskAbstract}{RGB}{245, 170, 185}  
\definecolor{taskPercept}{RGB}{180, 225, 200}   
\definecolor{taskPhysics}{RGB}{30, 60, 120}

\definecolor{resourceblue}{HTML}{4C7DBF}
\newcommand{\pending}[1]{\textcolor{red}{[#1]}}

\usepackage{xspace}

\usepackage[pr]{icml2026}

\makeatletter
\renewcommand{\pr@headerlogo}{%
  {\pr@logofont\textcolor{pr@accent}{%
    \href{https://solve-vision-with-code.com/}{solve-vision-with-code.com}}}%
}
\gdef\affil@sep{\enspace}
\newcommand{\affilbreak}{\gdef\affil@sep{\\[2pt]}}

\renewcommand{\printAffiliationsAndNotice}[1]{\global\icml@noticeprintedtrue%
  \stepcounter{@affiliationcounter}%
  \noindent\begin{minipage}{\textwidth}%
    \centering
    \fontsize{9}{11}\selectfont%
    \forloop{@affilnum}{1}{\value{@affilnum} < \value{@affiliationcounter}}{%
      \textsuperscript{\arabic{@affilnum}}\,\csname @affilname\the@affilnum\endcsname%
      \affil@sep\gdef\affil@sep{\enspace}%
    }%
    \ificmlshowauthors #1\fi%
    \ifdefined\icmlcorrespondingauthor@text
      \\[4pt]%
      \textsuperscript{\textdagger}\,Correspondence to: \icmlcorrespondingauthor@text.%
    \fi%
    \\[6pt]
  \end{minipage}%
  \vskip 8pt%
  {\centering\textcolor{pr@lightgray}{\rule{0.2\textwidth}{0.3pt}}\par}%
}
\makeatother

\usepackage{amsmath}
\usepackage{amssymb}
\usepackage{mathtools}
\usepackage{amsthm}

\usepackage[capitalize,noabbrev]{cleveref} 

\theoremstyle{plain}

\theoremstyle{definition}

\theoremstyle{remark}

\usepackage[disable,textsize=tiny]{todonotes}
\usepackage{enumitem}

\usepackage{makecell}
\usepackage{verbatim}   

\icmltitlerunning{Solve Vision with Code}

\begin{document}
\twocolumn[
  \icmltitle{Solve Vision with Code: Coding Agents Are Stronger Visual Reasoners Than Video Models}




\begin{icmlauthorlist}
    \icmlauthor{Dezhi Luo}{ucl}
    \icmlauthor{Fengyuan Yu}{ntu}
    \icmlauthor{Haotian Zhang}{usc}
    \icmlauthor{Zehong Zhao}{ucsd}
    \icmlauthor{Ran Ji}{ucsd}
    \icmlauthor{Siyuan An}
    {cmu}
    \icmlauthor{Huayi Qin}{ntu}
    \icmlauthor{Kaia Gao}{jhu}
    \icmlauthor{Zhongang Cai}{ntu}
    \icmlauthor{Lei Yang}{cuhk}
    \icmlauthor{Ziqi Huang}{ntu}
    \icmlauthor{Ruisi Wang}{ntu}
    \icmlauthor{Mark Wang}{ucb}
    \icmlauthor{Hokin Deng}{cmu}
\end{icmlauthorlist}

\icmlaffiliation{ntu}{Nanyang Technological University}
\icmlaffiliation{cmu}{Carnegie Mellon University}
\icmlaffiliation{cuhk}{Chinese University of Hong Kong}
\icmlaffiliation{ucsd}{University of California, San Diego}
\icmlaffiliation{ucb}{University of California, Berkeley}
\icmlaffiliation{jhu}{Johns Hopkins University}
\icmlaffiliation{ucl}{University College London}
\icmlaffiliation{usc}{University of Southern California}

\icmlcorrespondingauthor{Dezhi Luo}{dezhi.luo.24@ucl.ac.uk}
\icmlcorrespondingauthor{Hokin Deng}{hokind@andrew.cmu.edu}
\printAffiliationsAndNotice{}

\icmlkeywords{Visual Reasoning, Coding Agents, Video Models, Program Synthesis}
]




\newcommand{\inputsection}[2]{\IfFileExists{#1.tex}{\input{#1}}{\section*{\pending{section #2 missing: \texttt{\detokenize{#1.tex}}}}}}

\begin{figure*}[h]\centering\makebox[\textwidth][c]{\includegraphics[width=1.05\textwidth]{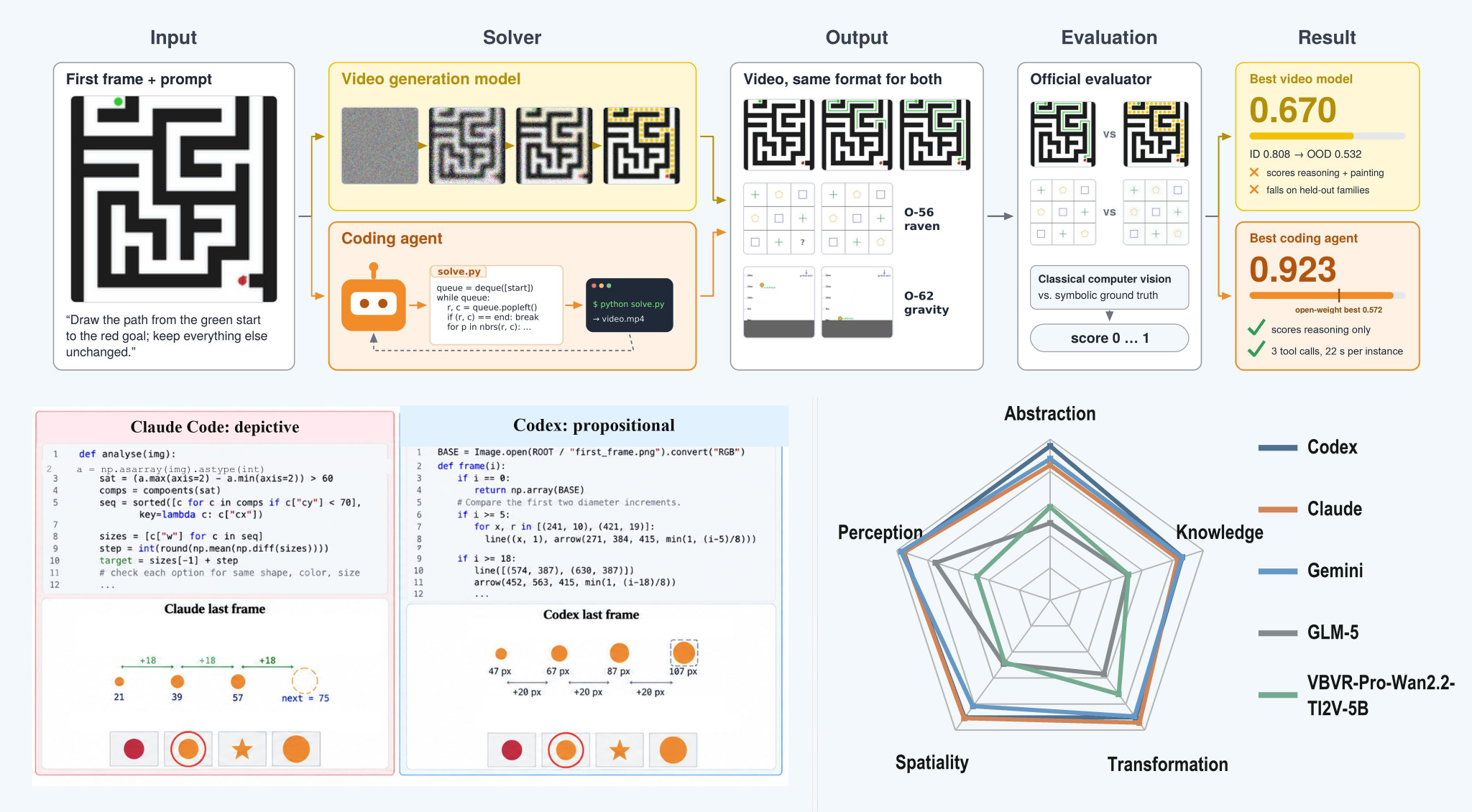}}\vspace{6pt}\caption{\textbf{Coding agents as visual reasoners.} Given the same first frame and prompt, a video generation model denoises the answer video directly, whereas a coding agent writes and executes a program that renders it. Both outputs share one video format and are scored by the same unmodified official evaluator, which compares them against symbolic ground truth using classical computer vision. The coding agent outperforms the best video model overall. The two solutions to one sequence task illustrate how the agents differ: Claude Code navigates the image to recover a transferable scheme (depictive), whereas Codex resolves the specific visual problem by hard-coding the answer it identified for that instance (propositional).}\label{fig:teaser}\end{figure*}

\begin{abstract}
Video generation models are increasingly evaluated as visual reasoners: given a first frame and a natural-language prompt, can the model produce the video in which the specified task is carried out? The paradigm is considered advantageous because the reasoning is native to the visual medium, unfolding directly in the frames rather than through language or code. Whether this native route is in fact the more reliable path to visual reasoning has not been tested against a symbolic alternative under the same conditions. We therefore pose the same tasks, with identical inputs, output specification and evaluator, to both media: rather than denoising pixels, a coding agent writes a program that renders the answer video, which an unmodified rule-based evaluator scores. On VBVR-Pro-Bench \citep{vbvrpro2026}, a 500-instance suite spanning five cognitive categories with symbolic ground truth, the agent receives the same first frame and prompt as the video model, with no access to ground truth, metadata or generation models, and one attempt per instance. Three closed-model agents (Codex with gpt-6-astra, Claude Code with Fable~5.1 and Gemini CLI with Gemini~3.1~Pro) score 0.923, 0.894 and 0.893, each more than 0.22 above the strongest video model. Among 23 runnable open-weight models, the best, GLM-5, reaches 0.572, above all video models except the RL-trained one. Together, we show that for tasks whose solutions can be computed from a first frame and a prompt, program synthesis is currently a substantially more reliable route to visual reasoning than video generation. We further distinguish depictive programs, which navigate the image to recover a transferable scheme, from propositional programs, which operate on hard-coded values, revealing how visual reasoning unfolds in agent-written programs and what it implies for its nature of reasoning in general.
\end{abstract}
\section{Introduction}
\label{sec:intro}

Video generation models now render photorealistic, temporally coherent footage across a wide range of domains \citep{openai2024sora, deepmind2025veo3, wan2025wan, kuaishou2025kling26, kong2024hunyuan}, prompting a growing line of inquiry into whether they can function as visual reasoners: producing videos that constitute solutions to structured tasks rather than renderings of descriptions \citep{wiedemer2025video,tong2026thinking, yang2025vrbench_maze, wang2026demystifying, newman2026video, v-reasonbench_luo2025v, ruler-bench_he2025ruler, mme-cof-pro_qi2026mme}. VBVR-Pro \citep{vbvrpro2026} provides a verifiable, large-scale standard for this question, pairing procedurally generated tasks with rule-based evaluation against symbolic ground truth. Each instance presents a first frame and a natural-language prompt, and the model must produce the video that carries out the prompt. VBVR-Pro-Bench comprises 100 task families with five instances each, and the evaluator scores the output frames across five cognitive categories: \emph{abstraction}, \emph{perception}, \emph{spatiality}, \emph{transformation} and \emph{knowledge}. Current video models remain substantially below the ceiling of this benchmark, including the strongest entry, a model RL-trained on the benchmark's task families with the evaluator as verifiable reward.

We address the same benchmark under a controlled change of medium, motivated by the observation that, for most VBVR-Pro tasks, the correct output is a deterministic function of the first frame and the prompt. A model that denoises pixels must recover this function implicitly within the generative process, whereas a model that writes code can state it explicitly and delegate rendering to an interpreter. We therefore replace the medium of the answer and hold every other element of the evaluation fixed. A coding agent receives the same first frame and prompt, together with the required container specification (frame size, frame rate, frame count and duration), and writes a program that renders the answer video. It has no access to the ground truth, the task generator or the task metadata, may not invoke an image or video generation model, runs in a CPU sandbox with standard Python imaging libraries under a 30-minute wall-clock budget, and has one attempt per instance. The output is scored by the unmodified official evaluator, and an instance without a delivered video scores zero, as it would for a video model that fails to generate (\cref{fig:teaser}). This instantiates the program-synthesis route to visual reasoning \citep{gao2023pal, chen2023pot, suris2023vipergpt, gupta2023visprog} on a benchmark designed for video models and assessed on their terms; related work is reviewed in \cref{sec:related}. We apply the protocol to all 500 instances with three closed-model coding agents (Codex with gpt-6-astra, Claude Code with Fable~5.1 and Gemini CLI with Gemini~3.1~Pro) and, through the OpenCode harness, 23 runnable open-weight models (DeepSeek~R1 is excluded as it does not support tool invocation on Amazon Bedrock). Video-model scores are taken from the published VBVR-Pro leaderboard and were not rerun.

Coding agents score substantially above video models. The three closed-model agents reach 0.923, 0.894 and 0.893 and lead in every task category, while the strongest video model, trained with reinforcement learning on the benchmark's task families with the evaluator as verifiable reward, reaches 0.670. The best open-weight model, GLM-5, scores 0.572, above every video model except the RL-trained one. Among open-weight models, low scores are driven mainly by delivery failure: models that never invoke a tool, stop before executing their code, or exhaust the time budget deliver few videos. Beyond the minimum tool use needed to deliver a video, the amount of tool use does not predict score, and the three closed agents, whose scores span a range of 0.030, differ by nearly an order of magnitude in cost. Finally, the agents' programs divide into two modes: depictive programs are general programs written by the agents that could solve any particularly instances of the task kind, whereas propositional programs encode an answer derived in the agent's session as symbolic constants and only render it, and by that such program written by the agent does not transfer the solving ability to another instance of the task kind. The depictive share is highest on \emph{perception} and \emph{transformation} tasks and the propositional share on \emph{abstraction} and \emph{knowledge} tasks, and neither mode predicts performance.

Together, we provide: (1) a \textbf{controlled comparison} in which coding agents answer VBVR-Pro tasks by writing a program that renders the answer video, under the same inputs, output container and official evaluator as the video models, covering three closed-model agents and 23 open-weight models against published video-model results on all 500 instances, with category-level scores; (2) an \textbf{efficiency analysis}, which separates delivery failure from content failure while accounting for tool calls, runtime and tokens; and (3) an analysis of the \textbf{reasoning strategies} behind the agents' programs, in which we classify the 1{,}385 programs with which the closed agents solve their own instance as \textit{depictive}, if they navigate the image at runtime to recover a general scheme that transfers to a different instance of the same task family, or as \textit{propositional}, if they operate on hard-coded values and are therefore confined to the single instance for which they were written, and find that the balance of the two strategies varies with the task category. These results collectively show that, for tasks whose solutions can be computed from a first frame and a prompt, explicit program synthesis currently outperforms native video generation by a substantial margin, and that the strategies by which agents reach this performance offer insight into how visual reasoning could be approached in world modelling, whether natively or through code.
\section{Method}
\label{sec:method}

This section describes the evaluation protocol: the agent and sandbox setup (\cref{sec:protocol}), the benchmark (\cref{sec:bench}), scoring (\cref{sec:scoring}) and efficiency accounting (\cref{sec:efficiency}). A coding agent receives the same first-frame and prompt inputs as a video model, writes a program that renders the answer video, and is scored by the benchmark's official evaluator. All results in this paper are produced by a single end-to-end pipeline that implements this protocol (\cref{sec:scoring}).

\subsection{Coding Agents as Visual Reasoners}
\label{sec:protocol}

\textbf{Task statement.} Each instance runs in a fresh container holding only its inputs together with a task statement generated by the harness. The statement is identical in structure for every agent and instance; only the instance-specific output format varies. It instructs the agent to write a program that renders the answer video and imposes three constraints: no image or video generation model may be called, only what the prompt requires may change, and the action must unfold over the full clip. 

\textbf{Sandbox.} The container is a CPU-only Python environment, standard imaging and scientific libraries, and a set of fonts preinstalled. The agent runs as an unprivileged user under fixed resource limits and a 30-minute wall-clock budget, with network access enabled for model API calls and package installation. The rules forbid generative image and video models, confirmed by an audit of every event log and retained program. Each instance receives exactly one attempt with no retry; an instance without a video at exit scores zero.

\textbf{Agents.} Each agent is an off-the-shelf command-line coding tool that drives a language model: it reads files, runs commands, edits code and iterates until it decides the task is complete. We run each tool in its official non-interactive mode with approvals bypassed and default settings, adding no system prompt, skill or other modification. \Cref{tab:agents} below lists the three closed-model agents along with the versions and models used respectively.

\subsection{The Benchmark}
\label{sec:bench}

The basic setup follows VBVR-Pro-Bench \citep{vbvrpro2026}. The input consists of a 1024×1024 image and a task prompt. The evaluation kit gives each family one or more of five category labels: \textit{Perception} 39 families, \textit{Abstraction} 25, \textit{Knowledge} 20, \textit{Transformation} 19, and \textit{Spatiality} 17, inherited from the original work. Sixteen families have more than one label. If a family has k labels, it counts as 1/k toward each of its categories, and a category's score is just the weighted average over the instances from those families.

\begin{table}[H]
\centering
\small
\setlength{\tabcolsep}{5pt}
\caption{The three closed-model coding agents, each the vendor's own CLI run in non-interactive mode with approvals bypassed. }
\label{tab:agents}
\begin{tabular}{lll}
\toprule
\textbf{Agent (CLI)} & \textbf{Version} & \textbf{Model} \\
\midrule
Codex CLI \citep{codexcli}     & 0.154.0 & gpt-6-astra \\
Claude Code \citep{claudecode} & 2.1.268 & Claude Fable 5.1 (Amazon Bedrock) \\
Gemini CLI \citep{geminicli}   & 0.59.0  & gemini-3.1-pro-preview \\
\bottomrule
\end{tabular}
\end{table}

\textbf{Open-weight models.} To test whether the findings are contingent to frontier closed models, we apply the same protocol with OpenCode \citep{opencode}, an open-source coding agent that drives any model through a standard chat API. We evaluate 24 open-weight models served by Amazon Bedrock from the GLM, Kimi, MiniMax, DeepSeek, Qwen, gpt-oss, Nemotron, Mistral, Llama and Gemma families, each under the same task statement, sandbox and time limit as the closed agents. One model, DeepSeek~R1, cannot call tools on Bedrock and is marked as unsupported, leaving 23 scored models.

\subsection{Scoring}
\label{sec:scoring}

Every video is scored by the official VBVR-Pro-Bench evaluation kit, unmodified, in the mode used for the published leaderboard. Each task family's rule-based evaluator compares the video with the symbolic ground truth using classical computer vision and OCR and returns a score in $[0,1]$. An instance without a video scores 0, as it does for a video model that fails to generate. A system's score is the mean over all 500 instances; category scores follow the weighting defined in \cref{sec:bench}. Video-model scores are those published with VBVR-Pro \citep{vbvrpro2026}, computed with the same kit; we do not rerun those models. 

\subsection{Efficiency Measures}
\label{sec:efficiency}
Each agent CLI writes a machine-readable log of its run, retained for every instance. From these logs we derive three quantities: \emph{tool calls} (actions the agent takes), \emph{tokens} (model input and output as reported by the agent) and \emph{wall-clock time} (elapsed time from container start to exit, including model latency). We report tool calls and tokens as means over 500 instances and wall-clock time as the median. We also record whether the agent called any tool, whether it reached the time limit, and whether a failed instance involved tool use; these flags support the failure analysis in \cref{sec:res-open}. Token counts are comparable within an agent and only indicative across agents, since tokenisers and caching differ.
\section{Results}
\label{sec:results}

\subsection{Main Results}
\label{sec:res-main}

\Cref{tab:main} reports the overall score of every system on the 500 instances of VBVR-Pro-Bench: the three closed-model coding agents, the 23 open-weight models that can be run as coding agents (DeepSeek~R1 excluded), and the 11 highest-scoring of the 14 video models on the published leaderboard \citep{vbvrpro2026}. All 37 systems are scored by the same unmodified evaluator; an instance without a video scores 0, as it would for a video model that fails to generate.


\begin{table}[h]
\centering
\caption{VBVR-Pro-Bench leaderboard, 100 task families $\times$ 5 instances, for coding agents and video models. All videos are scored by the unmodified official evaluator; the score is the mean over all 500 instances, and an instance without a video scores 0. Coding-agent rows are our runs (one attempt per instance, sandboxed, no image or video generation model); video-model rows are the published VBVR-Pro leaderboard numbers \citep{vbvrpro2026}, not rerun. \emph{Rank} is the position among all 37 systems. \emph{Videos} is the number of instances, out of 500, for which the agent wrote a video (--- for video models, which we did not run). The best coding agent and the best video model are shaded. $^\dagger$No video in any attempt (the model never calls a tool).}

\label{tab:main}
\footnotesize
\setlength{\tabcolsep}{4pt}
\begin{tabular}{rlcc@{\hspace{2em}}rlcc}
\toprule
Rank & System & Score & Videos & Rank & System & Score & Videos \\
\midrule
\multicolumn{4}{l}{\emph{Coding agents, closed models}} & \multicolumn{4}{l}{\emph{Coding agents, open-weight models (cont.)}} \\[1pt]
\cellcolor{bestcolor}1 & \cellcolor{bestcolor}Codex (gpt-6-astra) & \cellcolor{bestcolor}\textbf{0.923} & \cellcolor{bestcolor}500 & 30 & Nemotron Nano 3 30B & 0.028 & 91 \\
2 & Claude Code (Fable 5.1) & 0.894 & 500 & 31 & Qwen3-VL-235B-A22B & 0.025 & 70 \\
3 & Gemini CLI (Gemini 3.1 Pro) & 0.893 & 500 & 32 & Qwen3-32B & 0.004 & 25 \\
\cmidrule(r){1-4}
\multicolumn{4}{l}{\emph{Coding agents, open-weight models (OpenCode)}} & 33 & Llama 4 Maverick & 0.001 & 7 \\[1pt]
5 & GLM-5 & 0.572 & 495 & 34 & Gemma 3 27B & 0.000 & 0$^\dagger$ \\
7 & Kimi K2.5 & 0.557 & 498 & 35 & Gemma 3 12B & 0.000 & 0$^\dagger$ \\
11 & MiniMax M2.5 & 0.451 & 488 & 36 & Magistral Small & 0.000 & 0$^\dagger$ \\
14 & DeepSeek V3.2 & 0.378 & 475 & 37 & Llama 3.3 70B & 0.000 & 0$^\dagger$ \\
\cmidrule(l){5-8}
15 & Devstral 2 123B & 0.350 & 494 & \multicolumn{4}{l}{\emph{Video models (VBVR-Pro leaderboard)}} \\[1pt]
16 & Kimi K2 Thinking & 0.313 & 406 & \cellcolor{secondcolor}4 & \cellcolor{secondcolor}VBVR-Pro-Wan2.2-I2V-A14B (RL) & \cellcolor{secondcolor}0.670 & \cellcolor{secondcolor}--- \\
18 & gpt-oss-120b & 0.270 & 380 & 6 & VBVR-Pro-Wan2.1-I2V-14B & 0.562 & --- \\
19 & Qwen3-Coder-Next & 0.264 & 498 & 8 & VBVR-Wan2.2 & 0.517 & --- \\
20 & Nemotron Super 3 120B & 0.200 & 436 & 9 & Seedance 2.0 & 0.499 & --- \\
22 & Qwen3-Coder-30B-A3B & 0.143 & 480 & 10 & VBVR-Pro-Wan2.2-TI2V-5B & 0.470 & --- \\
23 & Mistral Large 3 675B & 0.127 & 254 & 12 & VBVR-Pro-LTX2.3 & 0.425 & --- \\
25 & GLM-4.7-Flash & 0.104 & 470 & 13 & Kling VIDEO 3.0 & 0.392 & --- \\
27 & gpt-oss-20b & 0.079 & 148 & 17 & Veo 3.1 & 0.309 & --- \\
28 & Qwen3-Next-80B-A3B & 0.070 & 167 & 21 & Wan2.2-I2V-A14B & 0.182 & --- \\
29 & Llama 4 Scout & 0.049 & 269 & 24 & LTX-2.3-I2AV & 0.112 & --- \\
 & & & & 26 & Wan2.1-I2V-14B-720P & 0.100 & --- \\
\bottomrule
\end{tabular}
\end{table}

\paragraph{Overall Ranking.} The three closed-model agents occupy the top three places, each delivering a video on all 500 instances (\cref{tab:main}). The best-performing video model, VBVR-Pro-Wan2.2-I2V-A14B, was trained with reinforcement learning on the in-domain task families using the benchmark's evaluators as reward \citep{vbvrpro2026}; it scores 0.670, a gap of 0.253 below Codex. The next-strongest video model, VBVR-Pro-Wan2.1-I2V-14B, is a further 0.108 behind.

\subsection{Results by Category}
\label{sec:res-category}

\Cref{tab:categories} reports each system's score on the benchmark's five task categories; a family with multiple labels contributes to each (\cref{sec:bench}). The closed-model agents lead in every category. Scores for each of the 100 task families are given in \cref{app:pertask}.

\begin{table}[h]
\centering
\caption{Score by task category for the three closed-model agents and the eight best open-weight models over all 500 instances, with the kit's label weighting (a family with $k$ labels counts $1/k$ toward each; number of families carrying each label in the second header row; best per column among the agents in bold, second underlined). The last row is the best video model: its category scores are VBVR-Pro's published per-split category scores \citep[Table 4]{vbvrpro2026} combined with the same weights (inputs rounded to three decimals), and its Overall is the published leaderboard score.}
\label{tab:categories}
\footnotesize
\setlength{\tabcolsep}{4pt}
\begin{tabular}{lcccccc}
\toprule
System & \textit{Abstraction} & \textit{Perception} & \textit{Spatiality} & \textit{Transformation} & \textit{Knowledge} & Overall \\
\emph{Families per label} & (25) & (39) & (17) & (19) & (20) & (100) \\
\midrule
\multicolumn{7}{l}{\emph{Coding agents, closed models}} \\[1pt]
Codex (gpt-6-astra) & \textbf{0.964} & \underline{0.942} & \textbf{0.903} & \underline{0.908} & \textbf{0.872} & \textbf{0.923} \\
Claude Code (Fable 5.1) & 0.828 & 0.934 & \underline{0.901} & \textbf{0.933} & \underline{0.851} & \underline{0.894} \\
Gemini CLI (Gemini 3.1 Pro) & \underline{0.887} & \textbf{0.960} & 0.816 & 0.884 & 0.850 & 0.893 \\
\midrule
\multicolumn{7}{l}{\emph{Coding agents, open-weight models (OpenCode)}} \\[1pt]
GLM-5 & 0.477 & 0.723 & 0.482 & 0.546 & 0.492 & 0.572 \\
Kimi K2.5 & 0.413 & 0.697 & 0.528 & 0.543 & 0.488 & 0.557 \\
MiniMax M2.5 & 0.351 & 0.602 & 0.427 & 0.403 & 0.334 & 0.451 \\
DeepSeek V3.2 & 0.308 & 0.521 & 0.368 & 0.277 & 0.276 & 0.378 \\
Devstral 2 123B & 0.289 & 0.465 & 0.375 & 0.282 & 0.234 & 0.350 \\
Kimi K2 Thinking & 0.233 & 0.453 & 0.348 & 0.201 & 0.195 & 0.313 \\
gpt-oss-120b & 0.224 & 0.344 & 0.295 & 0.166 & 0.245 & 0.270 \\
Qwen3-Coder-Next & 0.256 & 0.368 & 0.199 & 0.176 & 0.209 & 0.264 \\
\midrule
\multicolumn{7}{l}{\emph{Best video model}} \\[1pt]
VBVR-Pro-Wan2.2-I2V-A14B (RL) & 0.741 & 0.488 & 0.771 & 0.840 & 0.699 & 0.670 \\
\bottomrule
\end{tabular}
\end{table}

The three closed-model agents score above 0.8 in every category, with Codex leading on \textit{Abstraction}, \textit{Knowledge} and \textit{Spatiality}, Gemini CLI on \textit{Perception} and Claude Code on \textit{Transformation}. VBVR-Pro reports per-half category scores for each video model \citep[Table~4]{vbvrpro2026}; combining the two halves with the kit's task weights places VBVR-Pro-Wan2.2-I2V-A14B on the same footing as the agents (last row of \cref{tab:categories}). All three agents outperform it in every category, with the margin smallest on \textit{Transformation} and largest on \textit{Perception}. This ordering mirrors the video model's own profile: it scores highest on \textit{Transformation} and lowest on \textit{Perception}, while the agents span a narrower range across all five categories. The \textit{Perception} margin should be interpreted with caution: these tasks largely require locating and measuring objects in the frame, operations (connected-component analysis, colour filtering, etc) that rule-based code can execute exactly, which inflates the apparent advantage. The small margin on \textit{Transformation} may reflect the converse: transforming objects over time is a capacity video models may acquire more naturally from video data than from explicit algorithmic construction.

The open-weight models show a different category profile. \textit{Perception} is their strongest category, and the gap to the closed-agent mean is smallest there and largest on \textit{Abstraction} (\cref{tab:categories}). Reading the first frame correctly is within reach of the strongest open-weight models; rule inference is where they fall furthest behind.

\subsection{Delivery Failure and Efficiency Analysis}
\label{sec:res-open}

The strongest open-weight agent outperforms every video model except the RL-trained one; GLM-5 under OpenCode ranks fifth among all 37 systems and Kimi~K2.5 seventh (\cref{tab:main}). Below these two, scores decline steeply, tracking the number of videos delivered (Spearman's $\rho=0.93$, $n=23$). We distinguish \emph{delivery failure}, in which no video is produced when the container exits, from \emph{content failure}, in which a video is delivered but receives a low score. Because an instance without a video scores 0, delivery failure is the primary driver of the lowest scores. Three patterns of delivery failure emerge among the open-weight models:

\begin{itemize}[leftmargin=*,itemsep=2pt,topsep=2pt]
\item \textbf{No tool use.} Gemma~3~27B, Gemma~3~12B, Llama~3.3~70B and Magistral~Small never call a tool across 500 attempts and respond in prose; some emit a tool call as literal text with placeholder code, executing nothing and writing no file.
\item \textbf{Premature termination.} Qwen3-32B and Llama~4 Maverick average fewer than three tool calls per instance and deliver 25 and 7 videos respectively. Qwen3-32B typically writes a code skeleton and exits without running it; Llama~4 Maverick delegates to a non-existent sub-agent type and stops.
\item \textbf{Timeouts.} Qwen3-Coder-Next reaches the 1{,}800\,s wall-clock limit on 93 instances---more than all other models combined---but having written a video before the limit on all but two of these, it still delivers 498 videos. Its score of 0.264 reflects content quality rather than delivery failure.
\end{itemize}

\begin{wrapfigure}{r}{0.45\textwidth}\centering\includegraphics[width=1.0\linewidth]{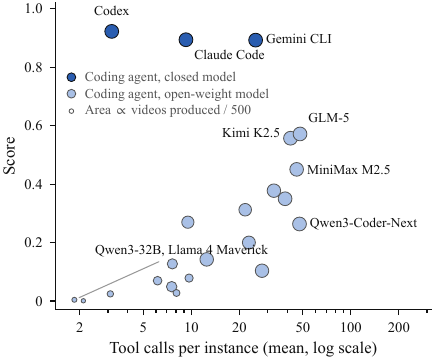}
\vspace{-4mm}
\caption{Score against tool calls per instance.}
\label{fig:tools}
\end{wrapfigure}

Among open-weight agents that produced a video, more tool calls accompany higher scores (Spearman's $\rho=0.88$; \cref{fig:tools}), because the lowest-scoring agents are those that terminate before delivering a video. This pattern does not extend to the upper range: the closed agents reach their scores with 3--25 calls per instance, whereas the two strongest open-weight agents make roughly 40--50 calls and more than twenty times as many input tokens as Codex, yet score lower by more than 0.3 (\cref{tab:efficiency}, with the full accounting in \cref{app:efficiency}). Beyond the minimum tool use required to deliver a video, call count does not predict score (as shown in Fig.\ref{fig:tools}).

\vspace{6mm}

The three closed agents score within 0.030 of one another but differ substantially in cost (\cref{tab:efficiency}). Relative to Codex, Claude Code uses roughly three times as many tool calls and four times as many input tokens, and Gemini CLI roughly eight times as many calls and twelve times as many tokens, for scores lower by only 0.029 and 0.030 respectively. Comparable task performance thus spans nearly an order of magnitude in cost; token counts are comparable within an agent but only indicative across agents, as tokenisers and caching differ (\cref{sec:efficiency}). Within an agent, more calls indicate harder instances. Partitioning each agent's 500 instances into quartiles by tool-call count, the highest-call quartile is the lowest-scoring for Claude Code (0.843) and Gemini CLI (0.819), consistent with agents probing, verifying and revising more on difficult instances. Call count is therefore better read as a marker of instance difficulty than of more thorough reasoning.  On the right (Fig. \ref{fig:tools} we show the score against mean tool calls per instance (log scale) for the 22 agents that produced at least one video. Disc area is proportional to the number of videos produced out of 500; the four agents with no video are omitted. Dark blue discs denote closed-model agents and light blue open-weight models under OpenCode. Values for the nine highest-scoring agents are given in \cref{tab:efficiency}.

\subsection{Distinct Strategies of Agentic Code-driven Visual Reasoning: Depictive and propositional programs}
\label{sec:strategy}

\begin{figure}[h]
\centering
\includegraphics[width=0.8\textwidth]{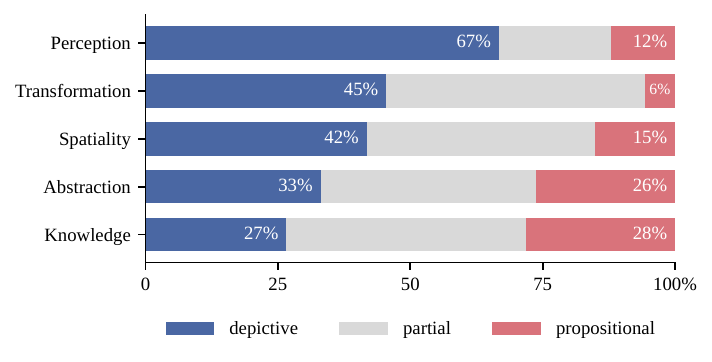}
\caption{Share of programs classified as depictive (blue) and propositional (red) by task category, for the closed-agent programs that solve their own instance. Classification is behavioural: each program is re-executed on a different instance of the same task family. A program is depictive if its answer follows the new input (it scores as well as before, or fails while measuring the new frame), and propositional if it runs but draws the same, now-incorrect answer, judged by the overlap of the region it changes in the final frame on the two instances. The grey band marks programs that fall between the two: they transfer only part of their answer, or draw a different but incorrect one. A task family that carries two category labels is split equally between them, and one with three labels is split three ways, as in \cref{tab:categories}.}
\label{fig:inversion}
\end{figure}

A coding agent's score does not reveal how its program arrives at the answer. We classified the 1{,}385 solve programs with which the three closed agents solve their own instance, and identified two reasoning strategies, distinguished by whether the program obtains the answer from the image or operates on values fixed in advance. A \emph{depictive} program navigates the pixel array of the first frame---filtering colours, measuring regions, locating objects---as one navigates a stored visual representation, and derives its answer from what it finds there. Its procedure is therefore tied to the task family rather than to the instance, and transfers to any instance of the same family. A \emph{propositional} program operates on hard-coded values---coordinates, colour values, cell indices---that the agent established beforehand and that the program only renders. It externalises the conclusion rather than the procedure that produced it, and is confined to the instance for which it was written: run on a different instance, it would render the same, now incorrect, answer. We assign the strategy behaviourally, by re-executing each program on a different instance of the same task family (\cref{fig:inversion}) among the four alternatives provided in VBVR-Pro-Bench \citep{vbvrpro2026}.

We name the two modes after the representational distinction at the centre of the mental imagery debate \citep{kosslyn1980image,pylyshyn1981imagery}. A depictive representation preserves the spatial structure of what it represents and supports reasoning by inspection---by depicting the properties from a stored array-like structure, as in early computational models of imagery \citep{kosslyn1977simulation}---whereas a propositional representation encodes information as discrete symbolic assertions in an amodal medium \citep{fodor1975language}, accessed by retrieval rather than spatial inspection. Depictive programs stand in the same relation to the first frame as hypothesised depictive imagery stands to a remembered scene: the first-frame is maintained as a spatial array, and the program reasons by navigating it. Propositional programs, conversely, encode the answer in a form that is indifferent to the spatial structure of the frame, stating what to render without re-examining the image. Depictive programs are themselves not analogue of the scene in any strict sense: they consist of symbolic code that navigates the arrays representing the image. Whether these programs' strategies correspond to processes in human visual cognition remains open, subjected to resolutions of the mental imagery debate, but we believe this diagnosis offers a helpful comparison for understanding how visual reasoning happens in agentic coding.

The balance of the two modes depends on the task category (\cref{fig:inversion}). Depictive programs account for 67\% of \textit{Perception} and 45\% of \textit{Transformation} programs---tasks that require locating, measuring and moving objects the code can find in the pixels. The share falls to 33\% for \textit{Abstraction} and 27\% for \textit{Knowledge}, where the propositional share is highest (26\% and 28\%, against 6\% for \textit{Transformation}); these tasks require recognising a rule or applying stored knowledge, and the agent more often settles the answer in the session. \textit{Spatiality} lies between the two, combining scene reading with path and relation computation (\cref{app:program-structure}). Among the closed agents, Claude Code keeps the measurement in the program more often (61\% depictive) than Codex (35\%). Gemini CLI opens the frame on only 10 of 500 instances (2.0\%), yet 17\% of its programs are propositional, the constants having been obtained by exploratory scripts during the session (\cref{app:fingerprint}). Neither mode signals performance: Codex has the highest score of the three despite only 24\% of its \textit{Abstraction} programs being depictive. The category alignment suggests that mode tracks the nature of the required reasoning: tasks that demand spatial operations on the frame favour depictive programs, whereas those that require rule inference or knowledge retrieval more often yield propositional ones.

\begin{figure}[h]
\centering
\includegraphics[width=0.9\textwidth]{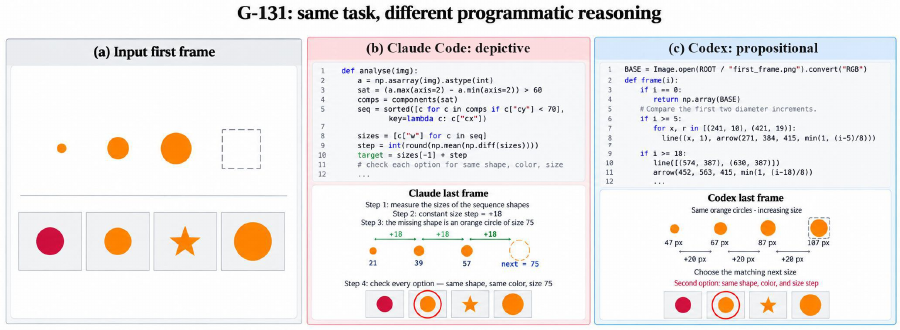}
\caption{Depictive and propositional programs on the same instance of G-131. (a) The input frame shared by both agents. (b) Claude Code (depictive): the program measures the circle widths, computes the constant difference and evaluates the candidates at run time. (c) Codex (propositional): the correct candidate is identified in the session, and the program only renders and highlights it. Both agents select the same candidate and obtain the same score.}
\label{fig:g131-reasoning}
\end{figure}

\Cref{fig:g131-reasoning} illustrates this separation on one instance of G-131, an increasing-size sequence task. The first frame shows three orange circles of increasing size in the first three positions of a row, an empty fourth position and four candidate answers. Claude Code and Codex select the same, correct candidate and obtain the same score, but their programs differ. Claude Code's program carries the whole procedure: it extracts the coloured regions, measures the circle widths (21, 39 and 57 pixels), computes their constant difference (18 pixels), predicts the next width (75 pixels) and tests each candidate against the recovered constraints on colour, shape and size. It is depictive and would recompute the answer on a different instance. Codex settles the answer during its session, and its program contains only fixed pixel coordinates and drawing operations that reproduce the explanation and highlight the chosen option. It is propositional and would highlight the same position on a different instance. Identical outputs on the instance at hand thus conceal programs of different generality, which is why score alone cannot identify the mode. Two further examples are given in \cref{app:examples}.

\section{Discussion}
\label{sec:discussion}

We evaluated coding agents on VBVR-Pro-Bench by having an agent write a program that renders the answer video from the same first frame and prompt that a video model receives, leaving the output format and the official evaluator unchanged. Under this protocol, all three closed-model agents clearly outperform the strongest video model, which was trained on the benchmark's own task families with the evaluator as reward, and the best open-weight agent also exceeds every other video model (\cref{sec:res-main,sec:res-open}). The agents' programs further reveal that the reasoning does not always take place in the program itself: in some cases the program recovers the answer from the frame when it runs (depictive programs), and in others the agent resolves the answer during its session and the program only renders it (propositional programs).

Taken together, the results indicate that coding agents currently solve the tasks of a benchmark built for video models more reliably than the video models themselves. This describes the present state of video reasoning and does not establish a limit in principle on what generative video models can learn. However, as an exploratory interpretation, it could suggest that native visual reasoning may be better suited to problems that depend less on symbolic, language-like operations. In humans, mental rotation, mechanical reasoning and intuitive physics are commonly attributed to visual-spatial simulation \citep{shepard1971rotation, hegarty2004mechanical, battaglia2013simulation}, whereas inferring a rule or applying stored knowledge is more naturally described in the discrete, compositional terms of language-of-thought accounts \citep{fodor1975language, quilty2023best}; how far mental imagery is itself depictive rather than propositional remains debated \citep{kosslyn1980image, pylyshyn1981imagery}. The division of reasoning modes by task category shows a similar correspondence in coding agents (\cref{sec:strategy}): tasks that call for locating, measuring and moving objects in the frame have the highest share of depictive programs, whereas tasks that require recognising a rule or applying stored knowledge have the highest share of propositional ones---the agent accesses the frame's information in the session rather than at run time and encodes the conclusion as symbolic constants in the program, separating inference from rendering. An interpretation of this is that coding agents, with their language backbone, may have inherent mechanistic properties that correspond to how humans navigate array-based mental imagery for reasoning, without instantiating the strongest depictive claim---that reasoning is itself embodied in the transformation of an analogue representation per se, one that is most intuitive for the conscious, flexible manipulation of mental imagery observed in humans. If so, while natural language supports reasoning in language models \citep{rothschild2025language}, most visibly through chain-of-thought \citep{wei2022chain}, the combination of code and visual information may serve as a less linguistically-demanding medium that navigates structured visual representations through symbolic operations, serving a similar function for visually grounded agents \citep{yang2025code}, so that ``native visual reasoning'' is not the exclusive province of systems that purely manipulate visual information (like video models), in accordance with some interpretations of human visual reasoning.

Several features of the benchmark and the comparison limit how far these findings generalise. Although VBVR-Pro-Bench was designed for video models, its tasks are generated and rendered by code, with flat colours, exact geometry and deterministic evaluators, a setting in which programmatic solutions are especially natural. The agents can decompose such scenes into colour masks, connected components and grid cells, in effect recovering the symbolic description from which the frame was rendered, and then produce the answer by editing pixels. This may not transfer to real-world scenarios in which objects are textured, occluded and variably lit, no symbolic ground truth exists, and an answer cannot be produced by editing a few regions of the first frame without a generative model. The comparison is also asymmetric: the agents use tools, a sandbox and up to 30 minutes of iteration, whereas the video models produce a single sample, and agent cost varies by more than an order of magnitude across models.

These limitations point to several directions for future work. Because the evaluator is deterministic, it can also serve as a reward, making a natural next step the training of the code-writing policy with GRPO \citep{shao2024deepseekmath}, mirroring the reinforcement learning that VBVR-Pro applied to video models \citep{vbvrpro2026}. The protocol could also be extended from a single first frame to video-to-video settings in three dimensions, such as WROP \citep{zhang2026training}, in which a model must continue a Blender-rendered clip while preserving object permanence and solidity; this would test whether the agents' advantage persists when the scene must be inferred from a clip rather than a single frame, although WROP is currently judged mainly by human preference, so a verifiable scorer would be needed. Beyond these, it remains open whether the advantage holds on real-world footage and whether hybrid systems that pair a coding agent with a generative renderer can combine the strengths of both. 
\section{Conclusion}
\label{sec:conclusion}

We evaluated coding agents on VBVR-Pro-Bench by having them write programs that render the answer video, with the benchmark's inputs, output format and official evaluator unchanged. Three closed-model agents score 0.893 to 0.923, against 0.670 for the strongest video model, which was trained with reinforcement learning on the benchmark's task families, and the best open-weight agent (0.572) exceeds every other video model. The agents' programs follow two strategies: depictive programs navigate the image to recover a general scheme that transfers to other instances of the same task family, whereas propositional programs operate on hard-coded values and are confined to the instance for which they were written. The depictive share is highest on tasks that require locating, measuring and moving objects found in the pixels, whereas the propositional share is highest on tasks that require recognising a rule or applying stored knowledge, where the agent more often derives the answer itself and encodes it in the program. Overall, these results show that explicit program synthesis currently outperforms native video generation by a substantial margin, and that, depending on the task, visual reasoning need not rely on systems that manipulate visual information alone: code operating on visual input can not only serve as a medium for abstract inference based on stored values, but may also support imagery-like reasoning by navigating the images descriptively when needed, implying an alternative approach to "native visual reasoning" based on a debate shared across human and machines.

\clearpage


\IfFileExists{main.bib}{%
  \bibliography{main}
  \bibliographystyle{icml2026}
}{\section*{\pending{main.bib missing: references not built}}}

\clearpage
\appendix
\onecolumn

\section{Related Work}
\label{sec:related}

\paragraph{Video models as visual reasoners.}

A growing body of work investigates whether video generation models can themselves serve as reasoners. \citet{wiedemer2025video} document zero-shot ``chain-of-frames'' behaviour in Veo~3 across perception, manipulation and maze tasks, establishing that models can express visual reasoning as generated frame sequences. \citet{tong2026thinking} frame video generation as a general multimodal reasoning paradigm and benchmark it across diverse tasks. Subsequent work probes where in the model this reasoning lives: \citet{wang2026demystifying} identify a ``chain-of-steps'' mechanism in which conclusions form over the denoising trajectory, and \citet{newman2026video} show that the commitment to a solution happens within the first few denoising steps. Several concurrent benchmarks measure the capability at scale: VR-Bench \citep{yang2025vrbench_maze} evaluates spatial reasoning through maze-solving and finds that fine-tuned video models outperform vision-language models on procedural spatial tasks; V-ReasonBench \citep{v-reasonbench_luo2025v}, RULER-Bench \citep{ruler-bench_he2025ruler} and MME-CoF-Pro \citep{mme-cof-pro_qi2026mme} cover a wider range of dimensions and reach a more cautious conclusion: current models are promising on short-horizon spatial coherence but unreliable on long-horizon causal reasoning and abstract logic. VBVR \citep{vbvr2026} and its successor VBVR-Pro \citep{vbvrpro2026} scale the question further. Hundreds of procedural generators produce first-frame, prompt and answer-video triples; a rule-based evaluator scores each task family; and VBVR-Pro-Bench holds 50 families out of all training to separate in-distribution performance from generalisation. VBVR-Pro additionally post-trains video models using the same evaluators as verifiable rewards under GRPO \citep{shao2024deepseekmath, guo2025deepseekr1} adapted to flow and diffusion architectures \citep{liu2026flow, xue2025dancegrpo}; its RL-trained checkpoints are the strongest video models on the benchmark. Most video generation benchmarks score with learned components. VBench \citep{huang2024vbench} mixes feature-based perceptual metrics; VideoScore \citep{he2024videoscore} trains a video-language judge; and LLM-as-a-judge protocols carry known positional and verbosity biases \citep{zheng2023judging}. VBVR-Pro instead evaluates with classical computer vision against symbolic ground truth, so the same video always receives the same score and the score can serve directly as a reward signal. This property makes the comparison in our paper exact: a coding agent and a video model are graded by the same deterministic function on the same 500 instances.

\paragraph{Code-driven visual reasoning.}

Writing a program has long been a route to visual understanding. Visual Programming \citep{gupta2023visprog} and ViperGPT \citep{suris2023vipergpt} prompt a language model to compose vision-module calls into executable Python; Visual Sketchpad \citep{hu2024visual} lets a multimodal model externalise intermediate results as drawn images. These systems produce text answers. On the language side, program-aided reasoning \citep{gao2023pal, chen2023pot} and Code as Policies \citep{liang2023code} showed that offloading computation to an interpreter improves reliability on arithmetic and embodied tasks. Off-the-shelf terminal coding agents---tools that iteratively read files, execute commands and revise programs in response to errors---are the current embodiment of this line of work. The evaluation pattern follows SWE-bench and SWE-agent \citep{jimenez2024swebench, yang2024swe}: one sandboxed container per instance, one attempt, and a verifier decides. A recent cluster of work uses code to produce visual content, in most cases without a generative video model in the loop. Kubrick \citep{he2024kubrick} coordinates director, programmer and reviewer agents that write Blender scripts to render text-described videos, and outperforms seven text-to-video baselines on FVD, FID, CLIP similarity and in a user study. VIGA \citep{yin2026vision} reconstructs reference images as executable 3D scenes through a render-and-inspect loop, supported by a purpose-built skill library and external asset generators, and simulates 4D physical dynamics through a physics engine. BVB \citep{tang2026bvb} targets video understanding: agents reconstruct real indoor videos as animated Blender scenes under a shared harness, scored by a VLM judge and a latent-similarity metric, and the best configuration retains only about half of the spatiotemporal facts in the source video despite high latent similarity. VideoCoCo \citep{li2026videococo} instead retains a video model: a coding agent writes a Blender program whose simulated draft conditions a generative video engine for photorealistic synthesis. Our setting differs from all four in three respects. First, the target is a verifiable reasoning benchmark in which a first frame and a prompt determine the answer, so correctness is decided by a deterministic evaluator rather than by distributional metrics, user studies or VLM judges. Second, the agents are off-the-shelf command-line tools used without modification, with no purpose-built skill library, asset retrieval or image or video generation model, and they render by writing pixels directly with standard imaging libraries rather than through a 3D engine. Third, the agents are scored on the same leaderboard as video models, including models trained on the benchmark itself, which permits a direct comparison of the two routes to the same task.

\section{Detailed Efficiency Analysis}
\label{app:efficiency}

\begin{table}[!htb]
\centering
\caption{Cost of solving an instance for the three closed-model agents and the six highest-scoring open-weight models. Values are means over the 500 instances computed from each agent's own event log, except run time, which is the median in seconds. Tool use is the fraction of instances in which the agent made at least one tool call, input and output tokens are given per instance in thousands, and timeouts are instances terminated at the 1{,}800\,s time limit; a video written before the limit is still scored. All agents that produced a video appear in \cref{fig:tools}. Claude Code's token counts cover the main model only; the auxiliary Sonnet~5 calls add 0.6\%.}
\label{tab:efficiency}
\small
\setlength{\tabcolsep}{4pt}
\begin{tabular}{lrrrrrrrr}
\toprule
Agent & Score & Videos & Tool use & Calls & Median s & In (k) & Out (k) & Timeouts \\
\midrule
\textbf{Codex (gpt-6-astra)} & \textbf{0.923} & \textbf{500} & \textbf{1.00} & \textbf{3.2} & \textbf{22} & \textbf{57} & \textbf{1.2} & \textbf{0} \\
Claude Code (Fable 5.1) & 0.894 & 500 & 1.00 & 9.2 & 105 & 233 & 7.3 & 0 \\
Gemini CLI (Gemini 3.1 Pro) & 0.893 & 500 & 1.00 & 25.3 & 218 & 659 & 7.5 & 0 \\
\midrule
GLM-5 & 0.572 & 495 & 1.00 & 48.0 & 378 & 1569 & 16.3 & 12 \\
Kimi K2.5 & 0.557 & 498 & 0.99 & 41.9 & 336 & 1213 & 19.2 & 0 \\
MiniMax M2.5 & 0.451 & 488 & 0.99 & 45.7 & 420 & 1295 & 19.8 & 0 \\
DeepSeek V3.2 & 0.378 & 475 & 1.00 & 32.9 & 277 & 636 & 13.6 & 0 \\
Devstral 2 123B & 0.350 & 494 & 0.99 & 38.7 & 344 & 839 & 13.1 & 0 \\
Kimi K2 Thinking & 0.313 & 406 & 0.96 & 21.8 & 125 & 509 & 15.5 & 1 \\
\bottomrule
\end{tabular}
\end{table}

\section{Per-Task Results of the Closed-Model Agents}
\label{app:pertask}

\Cref{tab:pertask} gives the score of each closed-model agent on every one of the 100 task families, with the family's split and cognitive category as the benchmark assigns  (each family counted under its first label, as in \cref{fig:inversion}; \cref{sec:bench} counts every label a family carries, so its totals exceed 100). The two halves are not balanced by category: the 50 in-domain families comprise 14 Knowledge, 13 Abstraction, 10 Spatiality, 7 Perception and 6 Transformation; the 50 out-of-domain families comprise 21 Perception, 12 Abstraction, 6 Spatiality, 6 Transformation and 5 Knowledge. On 32 of the 100 families all three agents score 1.000 (rounded to three decimal places) on all five instances. The lowest three-agent means are on G-43 (\texttt{understand\_scene\_structure}: 0.215, 0.003 and 0.770 for Codex, Claude Code and Gemini CLI), O-14 (\texttt{shape\_scale\_then\_outline}: 0.779, 0.008 and 0.576) and G-29 (\texttt{chart\_extreme\_with\_data}: 0.586, 0.566 and 0.320).

\begingroup
\scriptsize
\setlength{\tabcolsep}{4pt}
\begin{longtable}{llllrrr}
\caption{Per-task scores of the three closed-model coding agents on VBVR-Pro-Bench. Each entry is the mean official-evaluator score over the five instances of the task, an instance without a video counts 0, and the best of the three is in bold; the overall means match Table~\ref{tab:main}. In-Domain tasks (50) have training data in VBVR-Pro; Out-of-Domain tasks (50) are held out from all training.}
\label{tab:pertask}\\
\toprule
Task & Generator & Split & Category & Codex & Claude Code & Gemini CLI \\
\midrule
\endfirsthead
\multicolumn{7}{l}{\textit{Table~\ref{tab:pertask} continued}}\\
\toprule
Task & Generator & Split & Category & Codex & Claude Code & Gemini CLI \\
\midrule
\endhead
\midrule
\multicolumn{7}{r}{\textit{continued on next page}}\\
\endfoot
\bottomrule
\endlastfoot
G-3 & \texttt{stable\_sort} & ID & Perception & \textbf{1.000} & \textbf{1.000} & \textbf{1.000} \\
G-5 & \texttt{multi\_object\_placement} & ID & Transformation & \textbf{1.000} & \textbf{1.000} & \textbf{1.000} \\
G-8 & \texttt{track\_object\_movement} & ID & Spatiality & \textbf{1.000} & \textbf{1.000} & \textbf{1.000} \\
G-9 & \texttt{identify\_objects\_in\_region} & ID & Perception & 0.898 & \textbf{0.938} & 0.627 \\
G-13 & \texttt{grid\_number\_sequence} & ID & Spatiality & \textbf{1.000} & \textbf{1.000} & \textbf{1.000} \\
G-15 & \texttt{grid\_avoid\_obstacles} & ID & Spatiality & 0.414 & \textbf{0.805} & 0.609 \\
G-16 & \texttt{grid\_go\_through\_block} & ID & Spatiality & \textbf{0.845} & 0.843 & 0.758 \\
G-18 & \texttt{grid\_shortest\_path} & ID & Spatiality & \textbf{1.000} & \textbf{1.000} & \textbf{1.000} \\
G-21 & \texttt{multiple\_occlusions\_vertical} & ID & Knowledge & \textbf{0.996} & \textbf{0.996} & \textbf{0.996} \\
G-25 & \texttt{seperate\_object\_spinning} & ID & Transformation & \textbf{0.999} & \textbf{0.999} & 0.857 \\
G-29 & \texttt{chart\_extreme\_with\_data} & ID & Knowledge & \textbf{0.586} & 0.566 & 0.320 \\
G-31 & \texttt{directed\_graph\_navigation} & ID & Spatiality & 0.987 & \textbf{0.989} & \textbf{0.989} \\
G-39 & \texttt{attention\_shift\_different} & ID & Transformation & 0.620 & \textbf{0.740} & 0.618 \\
G-41 & \texttt{grid\_highest\_cost} & ID & Knowledge & \textbf{1.000} & 0.770 & \textbf{1.000} \\
G-43 & \texttt{understand\_scene\_structure} & ID & Knowledge & 0.215 & 0.003 & \textbf{0.770} \\
G-45 & \texttt{key\_door\_matching} & ID & Spatiality & \textbf{1.000} & \textbf{1.000} & \textbf{1.000} \\
G-51 & \texttt{predict\_next\_color} & ID & Abstraction & 0.926 & 0.934 & \textbf{0.995} \\
G-131 & \texttt{select\_next\_figure\_increasing\_size\_sequence} & ID & Abstraction & \textbf{1.000} & 0.799 & 0.786 \\
G-134 & \texttt{select\_next\_figure\_large\_small\_alternating\_sequence} & ID & Abstraction & \textbf{1.000} & \textbf{1.000} & 0.971 \\
G-138 & \texttt{spot\_unique\_non\_repeated\_color} & ID & Perception & 0.804 & \textbf{0.883} & 0.874 \\
G-158 & \texttt{identify\_all\_hollow\_points} & ID & Perception & \textbf{1.000} & \textbf{1.000} & \textbf{1.000} \\
G-194 & \texttt{construct\_concentric\_ring} & ID & Transformation & \textbf{1.000} & \textbf{1.000} & \textbf{1.000} \\
O-10 & \texttt{shape\_outline\_fill} & ID & Abstraction & \textbf{0.998} & 0.977 & 0.794 \\
O-12 & \texttt{shape\_color\_then\_scale} & ID & Abstraction & \textbf{0.970} & 0.969 & 0.964 \\
O-13 & \texttt{shape\_outline\_then\_move} & ID & Abstraction & \textbf{0.941} & 0.011 & 0.591 \\
O-14 & \texttt{shape\_scale\_then\_outline} & ID & Abstraction & \textbf{0.779} & 0.008 & 0.576 \\
O-15 & \texttt{ball\_bounces\_given\_time} & ID & Knowledge & \textbf{0.675} & 0.547 & 0.542 \\
O-16 & \texttt{color\_addition} & ID & Knowledge & \textbf{1.000} & \textbf{1.000} & \textbf{1.000} \\
O-18 & \texttt{glass\_refraction} & ID & Knowledge & \textbf{1.000} & 0.999 & 0.995 \\
O-19 & \texttt{mirror\_reflection} & ID & Knowledge & 0.765 & \textbf{0.840} & 0.797 \\
O-21 & \texttt{construction\_blueprint} & ID & Spatiality & \textbf{0.798} & 0.619 & 0.243 \\
O-23 & \texttt{domino\_chain\_branch\_path\_prediction} & ID & Knowledge & \textbf{0.833} & 0.482 & 0.588 \\
O-24 & \texttt{domino\_chain\_gap\_analysis} & ID & Knowledge & 0.758 & \textbf{0.886} & 0.758 \\
O-25 & \texttt{LEGO\_construction\_assembly} & ID & Spatiality & \textbf{1.000} & 0.852 & 0.662 \\
O-29 & \texttt{ballcolor} & ID & Abstraction & \textbf{0.728} & 0.411 & 0.487 \\
O-30 & \texttt{bookshelf} & ID & Abstraction & \textbf{0.938} & 0.923 & \textbf{0.938} \\
O-31 & \texttt{ball\_eating} & ID & Abstraction & 0.793 & \textbf{0.890} & 0.629 \\
O-32 & \texttt{rolling\_ball} & ID & Transformation & 0.324 & \textbf{0.697} & 0.612 \\
O-33 & \texttt{counting\_object} & ID & Perception & \textbf{0.935} & 0.693 & 0.872 \\
O-34 & \texttt{dot\_to\_dot\_task} & ID & Knowledge & 0.970 & 0.970 & \textbf{0.980} \\
O-36 & \texttt{grid\_shift} & ID & Transformation & 0.993 & 0.996 & \textbf{0.999} \\
O-37 & \texttt{light\_sequence} & ID & Perception & \textbf{1.000} & \textbf{1.000} & \textbf{1.000} \\
O-38 & \texttt{majority\_color} & ID & Perception & \textbf{1.000} & \textbf{1.000} & \textbf{1.000} \\
O-44 & \texttt{rotation\_puzzle} & ID & Abstraction & 0.990 & \textbf{0.993} & 0.893 \\
O-45 & \texttt{sequence\_completion} & ID & Abstraction & \textbf{1.000} & \textbf{1.000} & \textbf{1.000} \\
O-47 & \texttt{sliding\_puzzle} & ID & Abstraction & \textbf{1.000} & \textbf{1.000} & \textbf{1.000} \\
O-52 & \texttt{traffic\_light} & ID & Knowledge & 0.799 & \textbf{0.876} & 0.835 \\
O-53 & \texttt{clock} & ID & Knowledge & 0.977 & \textbf{0.983} & 0.980 \\
O-55 & \texttt{rotation} & ID & Spatiality & \textbf{0.681} & 0.635 & 0.412 \\
O-75 & \texttt{communicating\_vessels} & ID & Knowledge & 0.992 & \textbf{1.000} & 0.999 \\

\midrule
G-24 & \texttt{separate\_objects\_no\_spin} & OOD & Transformation & \textbf{0.964} & \textbf{0.964} & \textbf{0.964} \\
G-47 & \texttt{multiple\_keys\_for\_one\_door} & OOD & Spatiality & \textbf{0.999} & \textbf{0.999} & \textbf{0.999} \\
G-54 & \texttt{connecting\_color} & OOD & Perception & \textbf{1.000} & \textbf{1.000} & \textbf{1.000} \\
G-135 & \texttt{select\_next\_figure\_small\_large\_alternating\_sequence} & OOD & Abstraction & \textbf{1.000} & \textbf{1.000} & \textbf{1.000} \\
G-136 & \texttt{locate\_point\_in\_overlapping\_area} & OOD & Perception & \textbf{1.000} & \textbf{1.000} & \textbf{1.000} \\
G-140 & \texttt{locate\_topmost\_unobscured\_figure} & OOD & Spatiality & 0.966 & 0.978 & \textbf{0.988} \\
G-147 & \texttt{identify\_unique\_figure\_in\_uniform\_set} & OOD & Perception & \textbf{1.000} & \textbf{1.000} & \textbf{1.000} \\
G-160 & \texttt{circle\_largest\_numerical\_value} & OOD & Knowledge & \textbf{1.000} & \textbf{1.000} & \textbf{1.000} \\
G-161 & \texttt{mark\_second\_largest\_shape} & OOD & Perception & \textbf{1.000} & \textbf{1.000} & \textbf{1.000} \\
G-167 & \texttt{select\_longest\_polygon\_side} & OOD & Perception & \textbf{1.000} & 0.800 & \textbf{1.000} \\
G-168 & \texttt{identify\_nearest\_to\_square\_rectangle} & OOD & Perception & \textbf{1.000} & \textbf{1.000} & 0.999 \\
G-169 & \texttt{locate\_intersection\_of\_segments} & OOD & Perception & 0.993 & \textbf{1.000} & 0.986 \\
G-174 & \texttt{arrange\_circles\_by\_circumference} & OOD & Perception & \textbf{1.000} & \textbf{1.000} & \textbf{1.000} \\
G-189 & \texttt{draw\_midpoint\_perpendicular\_line} & OOD & Perception & 0.986 & \textbf{0.992} & 0.978 \\
G-193 & \texttt{draw\_next\_sized\_shape} & OOD & Abstraction & \textbf{0.989} & \textbf{0.989} & 0.987 \\
G-202 & \texttt{mark\_wave\_peaks} & OOD & Perception & \textbf{1.000} & \textbf{1.000} & \textbf{1.000} \\
G-206 & \texttt{identify\_pentagons} & OOD & Perception & \textbf{1.000} & \textbf{1.000} & \textbf{1.000} \\
G-212 & \texttt{find\_incorrect\_arrow\_direction} & OOD & Perception & \textbf{1.000} & \textbf{1.000} & \textbf{1.000} \\
G-217 & \texttt{circle\_central\_dot} & OOD & Perception & \textbf{1.000} & \textbf{1.000} & \textbf{1.000} \\
G-218 & \texttt{identify\_largest\_angle\_in\_triangle} & OOD & Perception & 0.200 & 0.600 & \textbf{1.000} \\
G-219 & \texttt{select\_leftmost\_shape} & OOD & Spatiality & \textbf{0.988} & \textbf{0.988} & 0.789 \\
G-221 & \texttt{outline\_innermost\_square} & OOD & Spatiality & 0.777 & \textbf{0.818} & 0.808 \\
G-222 & \texttt{mark\_tangent\_point\_of\_circles} & OOD & Perception & \textbf{1.000} & \textbf{1.000} & \textbf{1.000} \\
G-223 & \texttt{highlight\_horizontal\_lines} & OOD & Perception & 0.988 & \textbf{1.000} & \textbf{1.000} \\
G-240 & \texttt{add\_borders\_to\_unbordered\_shapes} & OOD & Perception & \textbf{0.946} & 0.930 & 0.905 \\
G-247 & \texttt{identify\_chinese\_character} & OOD & Knowledge & \textbf{1.000} & \textbf{1.000} & \textbf{1.000} \\
G-248 & \texttt{mark\_asymmetrical\_shape} & OOD & Perception & \textbf{1.000} & \textbf{1.000} & \textbf{1.000} \\
G-250 & \texttt{color\_triple\_intersection\_red} & OOD & Perception & 0.982 & 0.972 & \textbf{0.983} \\
G-273 & \texttt{high\_density\_liquid} & OOD & Knowledge & 0.905 & \textbf{0.935} & 0.900 \\
O-2 & \texttt{pigment\_color\_mixing\_subtractive} & OOD & Knowledge & \textbf{1.000} & \textbf{1.000} & \textbf{1.000} \\
O-5 & \texttt{symbol\_deletion} & OOD & Abstraction & \textbf{1.000} & 0.606 & \textbf{1.000} \\
O-6 & \texttt{2d\_geometric\_transformation} & OOD & Transformation & \textbf{0.970} & \textbf{0.970} & 0.842 \\
O-9 & \texttt{shape\_scaling} & OOD & Abstraction & \textbf{0.990} & 0.982 & 0.988 \\
O-11 & \texttt{shape\_color\_then\_move} & OOD & Abstraction & \textbf{0.998} & 0.732 & 0.991 \\
O-22 & \texttt{construction\_stack} & OOD & Abstraction & \textbf{0.983} & 0.850 & 0.745 \\
O-27 & \texttt{move\_2\_object\_to\_2\_target} & OOD & Transformation & \textbf{1.000} & \textbf{1.000} & 0.948 \\
O-39 & \texttt{maze} & OOD & Spatiality & \textbf{1.000} & \textbf{1.000} & \textbf{1.000} \\
O-43 & \texttt{object\_subtraction} & OOD & Perception & 0.949 & \textbf{0.981} & \textbf{0.981} \\
O-46 & \texttt{shape\_sorter} & OOD & Transformation & 0.998 & \textbf{0.999} & 0.949 \\
O-49 & \texttt{symmetry\_completion} & OOD & Spatiality & \textbf{1.000} & \textbf{1.000} & 0.800 \\
O-54 & \texttt{control\_panel} & OOD & Abstraction & \textbf{0.949} & 0.870 & 0.837 \\
O-56 & \texttt{raven} & OOD & Abstraction & \textbf{1.000} & \textbf{1.000} & 0.935 \\
O-58 & \texttt{symbol\_delete} & OOD & Abstraction & \textbf{1.000} & \textbf{1.000} & \textbf{1.000} \\
O-59 & \texttt{symbol\_insert} & OOD & Abstraction & 0.960 & \textbf{1.000} & 0.936 \\
O-60 & \texttt{symbol\_substitute} & OOD & Abstraction & \textbf{1.000} & \textbf{1.000} & \textbf{1.000} \\
O-61 & \texttt{symbol\_edit} & OOD & Abstraction & \textbf{1.000} & \textbf{1.000} & 0.896 \\
O-62 & \texttt{gravity\_physics} & OOD & Knowledge & 0.924 & \textbf{1.000} & 0.905 \\
O-64 & \texttt{animal\_matching} & OOD & Transformation & \textbf{0.991} & \textbf{0.991} & 0.982 \\
O-65 & \texttt{animal\_size\_sorting} & OOD & Perception & \textbf{1.000} & \textbf{1.000} & \textbf{1.000} \\
O-85 & \texttt{2d\_object\_rotation} & OOD & Transformation & 0.960 & \textbf{0.962} & 0.961 \\

\midrule
\multicolumn{4}{l}{\textit{Overall mean (100 tasks)}} & \textbf{0.923} & 0.894 & 0.893 \\
\end{longtable}
\endgroup


\begin{figure}[b]
\centering
\includegraphics[width=\textwidth]{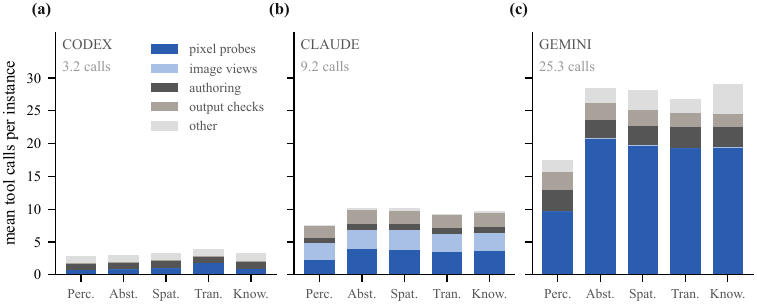}
\caption{Mean tool calls per instance, decomposed by capability, for the three closed agents across the five task categories. Each panel corresponds to one agent, with stacked bars partitioning its tool calls into pixel probing, image viewing, code authoring, output checking and all other operations. The within-agent composition of tool use changes relatively little across task categories, whereas the overall interaction profiles differ substantially between agents, forming distinct behavioural fingerprints.}
\label{fig:fingerprint}
\end{figure}

\section{Detailed Behavioural Analysis of the Closed-Model Agents}
\label{app:behaviour}

This appendix presents supplementary analyses of the closed agents' interaction traces and of the structural properties of the programs they generate. These analyses characterise how the three agents differ in their observable behaviour and working style. They do not bear directly on the main comparison between coding agents and video models, and they are placed here so as not to disrupt the flow of the primary results.

\subsection{Tool-Use Profiles}
\label{app:fingerprint}

Although the three closed agents face identical tasks under identical sandbox conditions, they follow markedly different solution paths. Codex tends to rely directly on its multimodal image input to interpret the first frame and typically begins writing code after only a small number of pixel-level probes. Claude Code pursues a more methodical approach, combining direct visual inspection with pixel-level measurement and frequently re-examining its own rendered output after generating the initial program. Gemini CLI rarely opens the image file directly; it instead reconstructs the content of the scene through scripts that programmatically inspect pixel values across the image, and it invokes OCR when the task requires extracting textual information.

To characterise these differences systematically, we classify every tool call in the 1{,}500 closed-agent traces into one of several capability categories: image viewing, pixel probing, OCR, code authoring, program rendering, output checking, package installation and environment setup. As \cref{fig:fingerprint} illustrates, each agent maintains a relatively stable distribution of tool use across the five task categories, while the three agents exhibit clearly distinct overall profiles. Quantitatively, Claude Code averages 2.8 image view operations and 3.3 pixel probe operations per task, and it verifies its own rendered output on 94.0\% of tasks; Codex performs self-verification on only 13.8\% of tasks. Gemini CLI averages 17.1 pixel probe operations per task while opening the image file directly on fewer than 2.0\% of tasks. Notably, all 53 package installation events observed across the three trace sets are performed by Gemini CLI, reflecting its systematic tendency to extend its information-processing capabilities through external libraries rather than through direct image access. Task content nonetheless modulates the activation of specific capabilities: OCR, for example, is invoked on 13.0\% of Knowledge tasks but on none of the Perception tasks. The overall tool-use composition thus functions primarily as a signature of agent-level behaviour, while the selective invocation of particular capabilities is additionally shaped by the demands of the specific task.

\subsection{Temporal Structure of Agent Sessions}
\label{app:session}

All three agents follow the same broad pattern of perceiving the task before committing to a program, but they differ substantially in the timing of the transition from information gathering to program construction and in the degree to which the two phases overlap. The median number of tool calls made before the first write to \texttt{solve.py} is 3 for Codex, 7 for Claude Code and 19 for Gemini CLI, indicating that what each agent considers sufficient information to begin implementation varies considerably across the three systems.

Dividing each session into early, middle and late phases reveals three qualitatively distinct temporal trajectories. Codex exhibits little evidence of discrete phase transitions: having acquired the information it judges necessary, it typically writes the complete program and submits the solution in a trajectory that is close to a single pass, with minimal iteration. Claude Code, by contrast, displays a pronounced inspect--author--verify structure: its early phase is dominated by viewing and measuring the task image, the middle phase by program construction, and the late phase by visual verification of its own rendered output, with 56\% of late-session tool calls consisting of image view operations. Gemini CLI shows no clear boundary between the perception and coding phases: pixel probing continues throughout the session, interleaved with successive program updates, and it performs on average approximately 3.4 code-writing or revision operations per task, yielding a probe-and-adjust trajectory that is iterative rather than staged. The three agents thus differ not only in the composition of their tool use but in the temporal organisation of information gathering, program construction and output verification, which persists consistently across task categories and instances.

\subsection{Program Structure by Category}
\label{app:program-structure}

The internal structure of the agents' programs varies systematically across task categories, complementing the distinction between depictive and propositional programs in \cref{sec:strategy}. In \textit{Spatiality}, 39\% of programs contain explicit combinatorial search---breadth-first search or shortest-path algorithms---indicating that a substantial fraction of spatial tasks reduce to discrete path-planning problems, which are more naturally expressed as search than as pixel manipulation. \textit{Knowledge} is the only category in which numerical physics simulation is prominent, with approximately 14\% of programs using discrete time steps, velocity components and restitution coefficients to model the required dynamics. \textit{Abstraction} exhibits the highest rate of direct pixel copying, at approximately 17\%: in several Raven-style tasks, programs copy pixel regions verbatim from one matrix cell to another, preserving the colour and stroke of the original rather than redrawing the shape procedurally. Together, these differences indicate that task categories correspond not only to different modes of reasoning, in the sense of \cref{sec:strategy}, but also to distinct computational strategies within the programs.

\subsection{Further Worked Examples}
\label{app:examples}

The comparison of the two modes on G-131 is given in \cref{sec:strategy}. This section presents two further examples.

\begin{figure}[t]
\centering
\includegraphics[width=\textwidth]{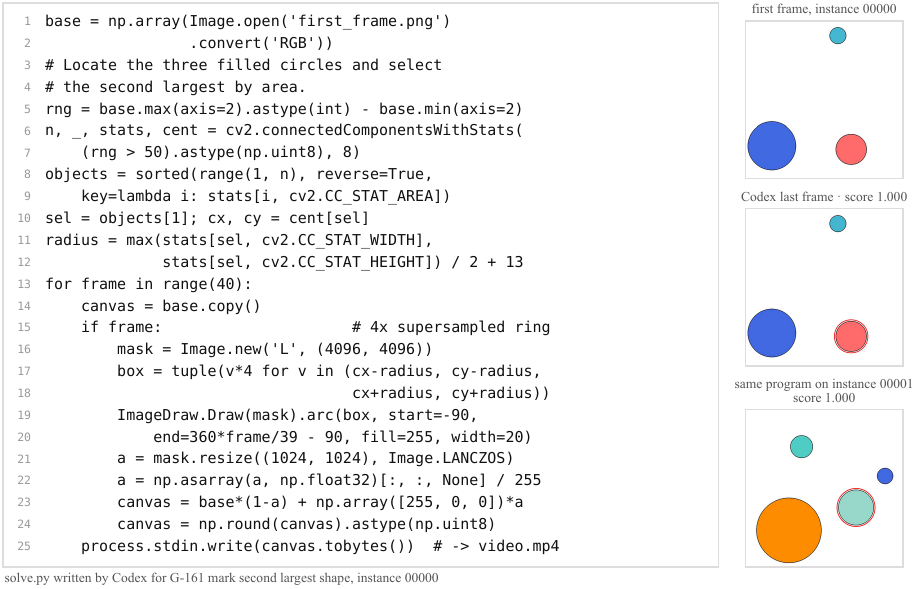}
\caption{The Codex program for G-161 (mark second largest shape, out-of-domain, instance 00000; 2 tool calls, score 1.0), trimmed to the lines discussed in the text, with the task's first frame, the last frame it renders, and the last frame the same program renders when re-executed on instance 00001, which has four circles instead of three.}
\label{lst:g161}
\end{figure}

\paragraph{G-161: a depictive program with run-time measurement.} \Cref{lst:g161} illustrates a depictive program on G-161, which asks for a red ring around the second largest of several circles. Codex receives the first frame as an image attachment and makes two tool calls, the first listing the directory and reading the prompt and the second writing and running the program. In its message it names the answer in prose (``the medium-sized circle at the lower right''), but the program makes no use of that sentence. It thresholds the frame on colour saturation, labels the connected components, sorts them by pixel area and takes the second (lines 5--10); the centre and radius of the ring come from that component's bounding box (lines 10--12). The only numbers written into the code are the saturation threshold, the ring's margin and width, and the frame count. Re-executed on instance 00001, which has four circles of other sizes, colours and positions, the program rings the second largest again and scores 1.0, although its own comment still speaks of ``three filled circles''. All 15 closed-agent programs for G-161 behave this way under the transfer test: 12 transfer with score 1.0 and 3 crash on the new frame; none draws its old answer. The run uses 780 output tokens.

\begin{figure}[!t]
\centering
\includegraphics[width=0.9\textwidth]{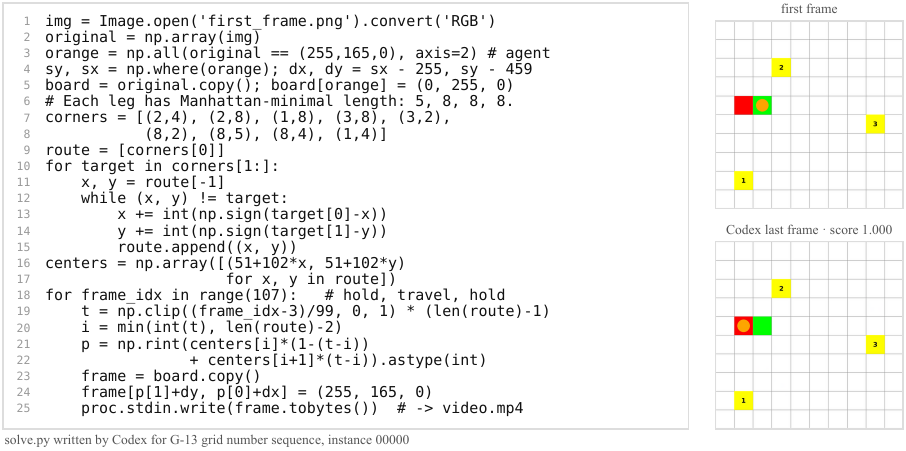}
\caption{The Codex program for G-13 (grid number sequence task, instance 00000; 3 tool calls, 20\,s, score 1.0), shown alongside the task's first frame and the final frame it renders. The nine corner cells of the route are written into the program as literals derived from the agent's visual reading of the image, with each leg annotated by its Manhattan length. The sprite is erased from its starting cell and re-stamped pixel-for-pixel at each interpolated position along the route, leaving the board itself unchanged throughout the video.}
\label{lst:grid}
\end{figure}

\paragraph{G-13: a propositional program with pixel-exact rendering.} \Cref{lst:grid} shows the Codex program for G-13, trimmed to the lines discussed here, a task that requires moving a sprite through a grid of numbered cells to the goal cell along a valid shortest path. Having probed the frame once to establish the bounding box of the sprite, Codex writes down the nine corner cells of the route directly from its visual reading of the image, annotating each leg with its Manhattan length as an inline comment, and then expands the sequence of corners into a full cell-by-cell trajectory. Each frame of the output video is constructed by taking the input image, erasing the sprite from its starting cell by restoring the background pixels, and re-stamping the sprite pixel-for-pixel at the interpolated position along the route, so that the grid itself---its numbers, lines and colouring---is never modified. The result is a video that is visually indistinguishable from one produced by a simulation of the physical task, obtained entirely by rendering the route the agent derived in its session.

\clearpage

\makeatletter
\setlength{\@fptop}{0pt}
\setlength{\@fpsep}{8pt plus 1fil}
\setlength{\@fpbot}{0pt plus 1fil}
\makeatother

\end{document}